\pdfoutput=1

\documentclass[11pt]{article}

\usepackage[final]{acl}

\usepackage{times}
\usepackage{latexsym}

\usepackage[T1]{fontenc}

\usepackage[utf8]{inputenc}

\usepackage{microtype}

\usepackage{inconsolata}

\usepackage{graphicx}
\usepackage{amsmath}
\usepackage{cleveref}
\usepackage{mathtools}
\usepackage{xcolor}
\usepackage{amsfonts}

\title{Promises, Outlooks and Challenges of Diffusion Language Modeling}

\author{Justin Deschenaux \\
  CLAIRE \\
  EPFL / 1015 Lausanne \\
  \texttt{justin.deschenaux@epfl.ch} \\\And
  Caglar Gulcehre \\
  CLAIRE \\
  EFPL / 1015 Lausanne \\
  \texttt{caglar.gulcehre@epfl.ch} \\}

\begin{document}
\maketitle
\begin{abstract}
  The modern autoregressive Large Language Models (LLMs) have achieved outstanding performance on NLP benchmarks, and they are deployed in the real world. However, they still suffer from limitations of the autoregressive training paradigm. For example, autoregressive token generation is notably slow and can be prone to \textit{exposure bias}. The diffusion-based language models were proposed as an alternative to autoregressive generation to address some of these limitations. We evaluate the recently proposed Score Entropy Discrete Diffusion (SEDD) approach and show it is a promising alternative to autoregressive generation but it has some short-comings too. We empirically demonstrate the advantages and challenges of SEDD, and observe that SEDD generally matches autoregressive models in perplexity and on benchmarks such as HellaSwag, Arc or WinoGrande. Additionally, we show that in terms of inference latency, SEDD can be up to 4.5$\times$ more efficient than GPT-2. While SEDD allows conditioning on tokens at abitrary positions, SEDD appears slightly weaker than GPT-2 for conditional generation given short prompts. Finally, we reproduced the main results from the original SEDD paper.
\end{abstract}

\section{Introduction}
\label{sec:intro}
Recent research on language models \citep{vaswani2017attention, devlin2018bert, radford2019language, brown2020language, kaplan2020scaling, raffel2020exploring, fedus2022switch, hoffmann2022training, chowdhery2023palm, team2023gemini, touvron2023llama}, particularly on autoregressive models, have produced systems with impressive capabilities for tasks such as coding \citep{chen2021evaluating}, math and reasoning \citep{trinh2024solving, romera2024mathematical, hosseini2024v}. 

In particular, \citet{romera2024mathematical} showed that Large Language Models (LLM) can provide novel solutions to hard problems in combinatorics. The solution of \citet{romera2024mathematical} relies on intensive sampling from an LLM to generate code modifications. In scenarios where a verifier can validate solutions, one bottleneck is the throughput of the LLM. Unfortunately, the autoregressive nature of current LLMs limits their generation speed. Various techniques have emerged to address sampling challenges and achieve faster generation. These include efficient implementations \citep{dao2022flashattention, kwon2023efficient, pope2023efficiently}, low precision inference \citep{dettmers2208llm, dettmers2023spqr, dettmers2023case}, novel architectures \citep{katharopoulos2020transformers, gu2021efficiently, gu2023mamba, poli2023hyena, orvieto2023resurrecting, peng2023rwkv} and multi-token predictions \citep{leviathan2023fast, chen2023accelerating, cai2024medusa, gloeckle2024better}.

Concurrently, there have been several parallel attempts at adapting diffusion models for text generation \citep{hoogeboom2021argmax, austin2021structured, li2022diffusion, han2022ssd, dieleman2022continuous, lou2023discrete, chen2023analog, gulrajani2024likelihood}. In this work, we compare the performance of autoregressive models against the approach of \citet{lou2023discrete}, called \textit{Score Entropy Discrete Diffusion} (SEDD). We believe that SEDD constitutes a viable alternative to autoregressive generation. For example, one key advantage of diffusion models is their ability to trade quality and compute flexibly. A reduced quality is acceptable if compensated by faster sampling in applications such as \citet{romera2024mathematical}, where a verifier is available.

The contributions of this work are as follows:
\begin{itemize}
  \item In \cref{sec:promises} and \cref{sec:challenges}, we describe the strengths of SEDD and promising research directions to improve SEDD.
  \item In \cref{sec:evaluations}, we reproduce the main results of \citet{lou2023discrete}, and include further evaluations on established benchmarks. Our findings suggest that SEDD performs comparatively with GPT-2 in unconditional text generation. However, SEDD appears slightly weaker than GPT-2 for conditional generation given short prompts.
\end{itemize}
\begin{table*}[t]
\centering
\begin{tabular}{|c|c|c|c|c|c|c|}
\hline
Model & \# params & LAMBADA & Wikitext2 & PTB &  WikiText103 & 1BW \\
\hline
GPT-2 (small) & 124M & \textbf{44.663} & 39.576 & 129.977 & 39.576 & \textbf{52.808} \\
\hline
SEDD (small) & 170M & $\leq$ 51.146 & \textbf{$\leq$ 38.941} & \textbf{$\leq$ 110.97} & $\leq$ \textbf{38.887} & $\leq$ 78.67 \\
\hline
\hline
GPT-2 (medium) & 255M & \textbf{35.879} & 29.575 & 116.516 & 29.575 & \textbf{43.627} \\
\hline
SEDD (medium) & 424M & $\leq$ 42.385 & $\leq$ \textbf{28.855} & $\leq$ \textbf{83.411} & $\leq$ \textbf{28.54} & $\leq$ 60.97 \\
\hline
\hline
GPT-2 (large) & 774M & 33.969 & 26.664 & 139.427 & 26.664 & 44.783 \\
\hline
\end{tabular}
\caption{\textbf{Zero-shot test perplexity}. For SEDD, we can only compute an upper bound. Our numbers differ slightly from \citet{lou2023discrete}. See \cref{sec:discrepancy-our-ppl-vs-sedd} for the explanation.}
\label{table:zero-shot-uncond}
\end{table*}
\section{Background}
\label{sec:background}
We succinctly describe the formalism from \citet{lou2023discrete}. Text generation consists of synthesizing sequences of length $L$, composed of elements from a vocabulary $\mathcal X \coloneq \{0, 1, ..., N - 1\}$. Furthermore, we assume access to training documents $\mathcal D := \{x_1, ..., x_m : x_i \in \mathcal X^L\}$. Ultimately, we want to sample from an unknown distribution $p_0 : \mathcal X^L \rightarrow [0, 1]$ such that $\sum_x p_0(x) = 1$. The distribution $p_0$ is assumed to have generated the samples in $\mathcal D$. As in continuous diffusion \citep{pmlr-v37-sohl-dickstein15, ho2020denoising, song2021scorebased}, we train a model to denoise corrupted examples and sample through iterative refinement starting from pure noise from a distribution $p_1$. 
\paragraph{Forward and reverse processes}
The continuous-time discrete diffusion process \citep{Anderson1991, campbell2022continuous, sun2023scorebased, benton2024denoisingdiffusionsdenoisingmarkov} is determined through $p_t : \mathcal X^L \rightarrow [0, 1]$, a family of distributions. The $p_t$ are defined through the differential equations $\frac{dp_t}{dt} = Q_t p_t$ (forward), and $\frac{dp_{T - t}}{p_{T - t}} = \bar Q_{T - t} p_{T - t}$ (reverse), where $\bar Q_t(x, y) = \frac{p_t(y)}{p_t(x)} Q_t (x, y)$ for $x \neq y$ and $\bar Q_t(x, x) = - \sum_{y \neq x} \bar Q_t (y, x)$. The distributions $p_0$ and $p_1$ represent the data and pure noise distributions. The forward and reverse processes depend on two objects. First, the transition operator $Q_t$ between pairs of sequences, a design choice, and second, the \textit{concrete score} $\frac{p(y)}{p(x)}$, approximated with a neural network. Since the domain $\mathcal X^L$ is large (e.g., $L = 1024$ and $N \approx 50$k), modeling transitions between all sequence pairs is impractical. \citet{lou2023discrete} side-steps this computational issue by corrupting each element of the sequence independently of each other, intuitively meaning that the forward process is defined with separate forward processes for each token. Additionally, the transition operator is chosen as $Q_t = \sigma_t Q$, with $Q \in \{ Q_\mathrm{uniform}, Q_\mathrm{absorb} \}$, and $\sigma_t \in [0, 1]$ representing a monotonically increasing noise schedule. Different choices of $Q$ implement different ways to corrupt tokens. When $Q = Q_\mathrm{uniform}$, tokens are replaced by randomly selected ones (with uniform distribution), while $Q = Q_\mathrm{absorb}$ means that tokens are replaced by a special MASK token, similar to BERT \cite{devlin2018bert}. Therefore, samples from $p_1$ are either sequences of random tokens or constant sequences of MASK.
\paragraph{Learning the concrete score} \citet{lou2023discrete} observed that score matching approaches \citep{JMLR:v6:hyvarinen05a, denoising-score-matching, meng2023concrete} were not effective to approximate $\frac{p(y)}{p(x)}$. As such, \citet{lou2023discrete} devised a novel loss, coined \textit{denoising score entropy}, minimized when the model learns the true concrete score. Conveniently, one can use the score entropy loss to compute a lower bound to the data likelihood. This bound is useful to compare SEDD with autoregressive models. The denoising score entropy loss $\mathcal L_\text{DSE}$ is defined as: 
\begin{center}
\small
\begin{equation*}
    \underset{\substack{x_0 \sim p_0\\ t \sim \mathcal U(0, 1]\\x \sim p_t(\cdot | x_0)}}{\mathbb{E}} \left[ \sum_{x \neq y} w_{xy} \left( s_\theta(x)_y - \frac{p(y | x_0)}{p(x | x_0)} \log s_\theta(x)_y \right)  \right]
    \label{eq:score-entropy}
\end{equation*}
\end{center}
where $s_\theta$ is a neural network that outputs arrays of shape $L \times N$ (like autoregressive language models). While the expression of $\mathcal L_\text{DSE}$ can initially look complicated, it is relatively easy to implement. Indeed, $\frac{p(y | x_0)}{p(x | x_0)}$ admits a closed-form expression. In practice, SEDD uses $w_{xy} = Q_t(x, y)$. Importantly, the weights $w_{xy}$ allow setting certain terms in the sum to zero. For example, when $Q = Q_\mathrm{absorb}$, \citet{lou2023discrete} sets $w_{xy}$ such that there is no loss on non-masked tokens.
\paragraph{Sampling} \citet{lou2023discrete} developed a sampling algorithm inspired by Tweedie's formula \citep{tweedie-formula}. However, one can also synthesize text using an Euler sampling strategy, by replacing the unknown concrete score with $s_\theta$ in the reverse process. For brevity, we direct the reader interested in the details to the work of \citet{lou2023discrete}.
%
%
%
%
%
\section{The promises of text diffusion}
\label{sec:promises}
\paragraph{SEDD and GPT-2 achieve similar test likelihoods}
SEDD was trained on the OpenWebText (OWT) dataset \citep{Gokaslan2019OpenWeb}, an open-source replication of the training data of GPT-2. As explained in \cref{sec:background}, one can approximate the likelihood of a dataset using $\mathcal L_\text{DSE}$. The data likelihood is a common metric to compare generative models \citep{theis2016note}. When evaluated on test datasets, SEDD matches or exceeds the likelihood of GPT-2, as seen in \cref{table:zero-shot-uncond}.
\paragraph{Sampling with fewer steps}
When sampling with SEDD, the number of forward passes and the sequence length are not tied. \citet{lou2023discrete} achieve better perplexity than sampling from GPT-2 without annealing with 32 sampling steps for sequences of 1024 tokens. Matching GPT-2 with nucleus sampling, however, requires 1024 steps. See \cref{table:gpt2-large-ppl} for details.
\paragraph{Simple sampling algorithm}
The sampling algorithm of \citet{lou2023discrete} is relatively straightforward. For those reasons, we believe that SEDD is a promising foundation for further research on sampling techniques. While differential equation samplers are viable, there might be simpler algorithms to synthesize text from SEDD. Identifying such methods would make diffusion-based language models more accessible to researchers without a STEM background.
\paragraph{Flexibility}
Unlike autoregressive models, SEDD does not generate tokens causally, and the forward process can be defined in various ways. \citet{lou2023discrete} obtained the best results with $Q = Q_\text{absorb}$, but the only constraint on $Q$ is to ensure that it is mass-preserving so that $p_t$ remains a valid distribution. Forward processes that enable repeated editions of tokens would be suited for tasks that requires reasoning over long sequences such as theorem proving to allow correcting previous mistakes. As such, designing better transition operators is a promising direction. 
\paragraph{A familiar architecture}
The transformer backbone of SEDD is similar to the diffusion transformer of \citet{peebles2023scalable} and has identical input and output dimensions as GPT-2. Therefore, one could instead use any sequence model, such as state-space models. Evaluating the trade-offs associated with different architectures is an open research direction.
\section{The challenges of text diffusion}
\label{sec:challenges}
One important challenge of diffusion language modeling is due to difficulty of implementing it correctly and tuning it properly when compared to simple autoregressive training. Below, we list a few other challenges that would improve the text quality of SEDD if addressed.
\paragraph{$Q_\text{absorb}$ only denoises once} 
In \cref{sec:promises}, we suggest that the ability to edit previously generated tokens is important for reasoning tasks. Unfortunately, \citet{lou2023discrete} achieve their best results using $Q = Q_\text{absorb}$. In particular, using $Q_\text{absorb}$ implies that a token is \textit{not} edited multiple times. We therefore encourage research in this direction.
\paragraph{KV-caching is non-trivial}
Unlike autoregressive models, SEDD cannot benefit from KV-caching because its attention computation is not causal. Indeed, when a new token is generated, it will influence the attention matrix of all tokens. Additionally, SEDD is conditioned on the noise level $\sigma_t$, complicating computation caching further.
\paragraph{Mask tokens waste FLOPs}
When using the $Q_\text{absorb}$ transition operator, a large fraction of the tokens processed by the model are masked and hence do not provide any information. This leads to wasted FLOPs. An architecture or a training procedure that does not require processing masked tokens would significantly improve the sampling speed.
\paragraph{Variable-size generation}
The fixed generation length of SEDD is a major obstacle for open-ended generation. Unlike autoregressive models that dynamically adjust the sequence length, SEDD operates on arrays of pre-defined size. This issue is connected to the previous one; we believe they might be fixed together.
\section{Experimental results}
\label{sec:evaluations}
\begin{table*}[t]
\caption{\textbf{Accuracy on lm-harness suite tasks} \citep{leo_gao_2021_5371629}. While on most tasks GPT-2 seems to perform better, the accuracies of SEDD and GPT-2 are close. See \cref{sec:lm-eval-harness-details} for more details.}

\centering
\begin{tabular}{|c|c|c|c|c|c|c|c|}
\hline
Model & \# params & LAMBADA & HellaSwag & PIQA &  Arc Easy & Arc Chall. & WinoGrande \\
\hline
GPT-2 (small) & 124M & 32.54 & 28.94 & \textbf{62.89} & \textbf{43.85} & \textbf{18.94} & \textbf{51.69} \\
\hline
SEDD (small) & 170M & \textbf{42.73} & \textbf{29.30} & 58.32 & 40.28 & 18.60 & 50.67 \\
\hline
\hline
GPT-2 (medium) & 355M & 43.06 & \textbf{33.26} & \textbf{67.62} & \textbf{49.11} & \textbf{21.58} & \textbf{52.95} \\
\hline
SEDD (medium) & 424M & \textbf{54.50} & 32.89 & 62.13 & 46.55 & 21.33 & 50.99 \\
\hline
\hline
GPT-2 (large) & 774M & 47.68 & 36.39 & 70.34 & 53.15 & 21.75 & 55.48 \\
\hline
\end{tabular}
\label{table:lm-eval-harness}
\end{table*}
We compare the quality, diversity and latency of SEDD and GPT-2.
\subsection{Unconditional generation quality}
\label{sec:uncond-quality}
We compare SEDD and GPT-2 based on their zero-shot perplexity on test datasets and based on the likelihood of generated text using a larger model (GPT-2 large, 770M parameters), similar to \citet{savinov2022stepunrolled, strudel2022self, dieleman2022continuous}. As observed by \citet{lou2023discrete}, SEDD produces text with lower perplexity than GPT-2 without annealing. When sampling with 1024 steps, SEDD produces text of similar likelihood as annealed sampling from GPT-2. See \cref{table:zero-shot-uncond,table:gpt2-large-ppl} for more details.
\subsection{Conditional generation quality}
\label{sec:cond-quality}
We evaluate the conditional generation quality using automated metrics and the lm-eval-harness suite of \citet{leo_gao_2021_5371629}.
\paragraph{Automated metrics}
We compare the quality of text generated by SEDD and GPT-2 using the MAUVE metric \citep{pillutla2021mauve} and the perplexity of sampled continuations using GPT-2 large. We compute those metrics using prefixes and continuations of 50 tokens each. We extract 1000 prompts from the \href{https://github.com/openai/gpt-2-output-dataset}{test set of OpenAI's webtext dataset}. While \citet{lou2023discrete} suggests that SEDD medium matches GPT-2 in quality, we observed that SEDD achieves slightly lower MAUVE score. Reproducing the MAUVE results of \citet{lou2023discrete} was non-trivial, and we observed that \textbf{generating more tokens achieves better quality than generating 50 only, even if we use 100 tokens to compute MAUVE}. Indeed, sampling the 50 tokens of the continuation only resulted in poorer performance. For more details on this surprising result and exact MAUVE values, refer to \cref{sec:reproduce-mauve-results}.
\paragraph{Downstream performance}
We compare the accuracy of SEDD and GPT-2 on downstream tasks using the evaluation suite of \citet{leo_gao_2021_5371629}. The results are shown in \cref{table:lm-eval-harness}. We noticed the performance of SEDD and GPT-2 overall are very close\footnote{For implementation details, refer to \cref{sec:lm-eval-harness-details}}.
\paragraph{Diversity}
We evaluate diversity based on statistics computed on the training data and synthesized samples. See \cref{tab:diversity} for all details. Surprisingly, the repetition rate of SEDD increases as the number of sampling steps increases. Similarly, the unigram entropy negatively correlates with the number of sampling steps. Importantly, SEDD appears less diverse than GPT-2 with and without annealing.
\subsection{Sampling speed}
\begin{figure}
    \centering
    \includegraphics[width=0.85\linewidth]{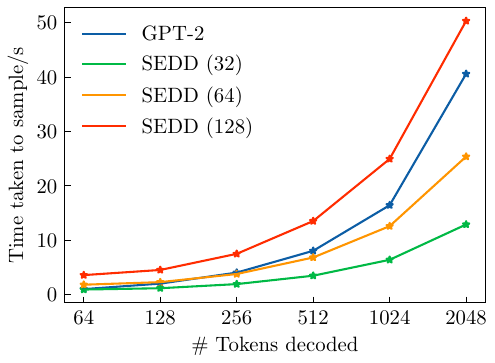}
    \caption{\textbf{Latency results during inference:} GPT-2 (1.3B, with KV-caching) vs. SEDD (1.45B, without caching). \textbf{SEDD (k)}: sampling from SEDD with k steps.}
    \label{fig:sampling-speed}
    \vspace{-3mm}
\end{figure}

We compare the generation latency of SEDD and GPT-2 with KV-caching on a single NVIDIA A100-SXM4-40GB GPU in \cref{fig:sampling-speed}. See \cref{sec:timing} for more details.
\section{Conclusion}
\label{sec:conclusion}
In this work, we argue that diffusion models for text are a relevant alternative to autoregressive generation, since SEDD achieves similar generation quality as GPT-2, while offering more flexibility at sampling. Additionally, we present the advantages and challenges of current SEDD models. Most notably, we believe that improving the sampling efficiency of SEDD is crucial to enable its applications. For example, matching the unconditional text quality of GPT-2 with nucleus sampling requires many steps and, hence, is significantly slower than GPT-2 with KV-caching. In parallel to the efficiency aspect, alternative definitions of the forward process operator are relevant for application on reasoning tasks. According to empirical analysis while we are hopeful for the future of diffusion language models, it is still early to say that they will dethrone autoregressive models in the immediate future.

\newpage
\section{Limitations}
\label{sec:limitations}
We present the main limitations of SEDD in \cref{sec:challenges}. We evaluate relatively small SEDD models (less than 500M parameters) because we are not aware of open-source text diffusion models in the billion-parameter range. It is uncertain how much the capabilities of SEDD would improve, given further computing, or whether it would still match autoregressive models. Additionally, we evaluated the checkpoints released by \citet{lou2023discrete} instead of training models from scratch. Finally, our evaluations rely on automated metrics such as the likelihood of test dataset, the likelihood of generated text using a large model, MAUVE, token-level statistics and specific benchmarks. Evaluation through direct human feedback would be more reliable to judge the quality, albeit more expensive.
\section{Ethical considerations}
This work focuses on SEDD, a recent model proposed by \citet{lou2023discrete}. Our primary goal is to present its promises and challenges to the NLP community, rather than introducing new techniques. Studying the capabilities and potential risks of language models is crucial. Note that the relative small size and lack of instruction-tuning of SEDD makes SEDD less concerning than modern large language models. This study cannot conclude on the capabilities of future larger text diffusion models.
\section{Acknowledgements}
We are grateful to Razvan Pascanu for insightful discussions and feedback on our work.
\newpage
\bibliography{custom}

\newpage
\appendix
\begin{table}[t]
\centering
\begin{tabular}{|c|c|}
\hline
Model/Source & Perplexity (GPT-2 large) \\
\hline
OpenWebText & 13.269 \\
\hline
\hline
GPT-2 (reg) & 140.134 \\
\hline
GPT-2 (top-p 0.95) & 42.691 \\
\hline
\hline
SEDD (100 steps) & 84.292 \\
\hline
SEDD (256 steps) & 63.301 \\
\hline
SEDD (512 steps) & 51.050 \\
\hline
SEDD (1024 steps) & \textbf{40.648} \\
\hline
\end{tabular}
\caption{Perplexity of 1000 samples generated by \textbf{small} (140-170M) models, evaluated using larger models.}
\label{table:gpt2-large-ppl}
\end{table}
\begin{table*}[t]
\centering
\begin{tabular}{|c|c|c|c|}
\hline
Model / Dataset & Fraction unique unigram  & Unigram H & Rep. rate (100)  \\
\hline
OWT & 0.49 ± 0.07 & 5.14 ± 0.31 & 0.033 ± 0.013 \\
\hline
\hline
GPT-2 (reg) & 0.59 ± 0.06 & 5.89 ± 0.24 & 0.030 ± 0.007 \\
\hline
GPT-2 (p=0.95) & 0.50 ± 0.08 & 5.56 ± 0.45 & 0.030 ± 0.008 \\
\hline
\hline
SEDD (100 steps) & 0.48 ± 0.03  & 5.54 ± 0.11 & 0.033 ± 0.005 \\
\hline
SEDD (256 steps) & 0.45 ± 0.04 & 5.45 ± 0.15 & 0.033 ± 0.005 \\
\hline
SEDD (512 steps) & 0.42 ± 0.04 & 5.36 ± 0.17 & 0.034 ± 0.006 \\
\hline
SEDD (1024 steps) & 0.39 ± 0.05 & 5.24 ± 0.23 & 0.034 ± 0.006 \\
\hline
\end{tabular}
\caption{Diversity metrics for 1000 sequences of 1024 tokens from the training data and small models. Observe that the fraction of unique unigrams decreases as the number of sampling steps increase. At the same time, the repetition rate (fraction of tokens that appear in the previous 100 tokens) remains stable and close to the training data. \textbf{Unigram H}: entropy, the distribution is computed by normalizing token counts by the sequence length. \textbf{GPT-2 (reg)}: sampling without annealing. \textbf{GPT-2 (p=0.95)}: nucleus sampling.}
\label{tab:diversity}
\end{table*}
\begin{table*}[t]
\centering
\begin{tabular}{|c|c|c|c|}
\hline
Model & MAUVE (gen 1024 tokens) & MAUVE (gen 100 tokens) & cont. PPL \\
\hline
GPT-2 (reg)         & 0.9768 &  \textit{0.9768} & 68.956 \\
\hline
GPT-2 (top p=0.95)  & \textbf{0.9843} & \textit{\textbf{0.9843}} & 41.383 \\
\hline
GPT-2 (top k=50)    & 0.9710 & \textit{0.9710} & \textbf{17.166} \\
\hline
\hline
SEDD (128 steps)  & 0.9422  & 0.9358 & 125.517 \\
\hline
SEDD (256 steps)  & 0.9579  & 0.9149 & 99.028 \\
\hline
SEDD (512 steps)  & 0.9619  & 0.9351 & 88.725 \\
\hline
SEDD (1024 steps) & 0.9441  & 0.9231 & 76.935 \\
\hline
\end{tabular}
\caption{Conditional generation quality of \textbf{small} models. MAUVE is computed using prompts and continuations of 50 tokens each. \textbf{MAUVE (gen 1024 tokens)}: MAUVE scores when sampling sequences of 1024 tokens and use the first 100 to compute MAUVE. \textbf{MAUVE (gen 100 tokens)}: MAUVE scores when sampling sequences of 100 tokens directly. Interestingly, MAUVE scores improve when sampling longer sequences. Note that SEDD was trained with sequences of 1024 tokens. Observe that GPT-2 seems to achieve better performance than SEDD. \textbf{cont. PPL}: perplexity of the first 50 tokens of the continuation (from a total of 1024 tokens).}
\label{table:cond-quality-mauve-ppl-small}
\end{table*}
\begin{table*}[t]
\centering
\begin{tabular}{|c|c|c|c|}
\hline
Model & MAUVE (gen 1024 tokens) & MAUVE (gen 100 tokens) & cont. PPL \\
\hline
GPT-2 (reg)        & 0.9788 & \textit{0.9788} & 42.861 \\
\hline
GPT-2 (top p=0.95) & \textbf{0.9808} & \textit{\textbf{0.9808}} & 26.503 \\
\hline
GPT-2 (top k=50)   & 0.9775 & \textit{0.9775} & \textbf{12.470} \\
\hline
\hline
SEDD (128 steps)  & 0.9645  & 0.9147 & 81.113 \\
\hline
SEDD (256 steps)  & 0.9611  & 0.9254 & 67.754 \\
\hline
SEDD (512 steps)  & 0.9688   & 0.9217 & 59.320 \\
\hline
SEDD (1024 steps) & 0.9583  & 0.9242 & 52.192 \\
\hline
\end{tabular}
\caption{Conditional generation quality of \textbf{medium} models. MAUVE is computed using prompts and continuations of 50 tokens each. \textbf{MAUVE (gen 1024 tokens)}: MAUVE scores when sampling sequences of 1024 tokens and use the first 100 to compute MAUVE. \textbf{MAUVE (gen 100 tokens)}: MAUVE scores when sampling sequences of 100 tokens directly. Interestingly, MAUVE scores improve when sampling longer sequences. Note that SEDD was trained with sequences of 1024 tokens. Observe that GPT-2 seems to achieve better performance than SEDD. \textbf{cont. PPL}: perplexity of the first 50 tokens of the continuation (from a total of 1024 tokens).}
\label{table:cond-quality-mauve-ppl-medium}
\end{table*}

\section{Appendix}
\subsection{Discrepancy in perplexity against the original SEDD paper}
\label{sec:discrepancy-our-ppl-vs-sedd}
The results reported in \cref{table:zero-shot-uncond} differ from the results reported by \citet{lou2023discrete}. The difference is explained by the distinct way we compute the perplexity over a dataset. We compute the perplexity by averaging the negative log-likelihood (loss) of the data, followed by exponentiation. On the other hand, \citet{lou2023discrete} first exponentiate the sequence-level loss before averaging. We selected this order of operation because it aligns with common practices in our experience. Concretely, assume that the array \verb|A| of length $N$ contains the sequence-level negative log-likelihood over the test dataset. \citet{lou2023discrete} compute the dataset-level perplexity as \verb|A.exp().mean()|, while we compute it as \verb|A.mean().exp()|. Note that the conclusions remain the same, as SEDD achieves better perplexity for the same datasets with both orders.
\subsection{Evaluation using lm-eval-harness}
\label{sec:lm-eval-harness-details}
To evaluate the capabilities of SEDD and GPT-2, we leverage the lm-eval-harness test suite developed by \citet{leo_gao_2021_5371629}. Their library offers a standardized interface to evaluate language models on a broad range of tasks. For instance, \citet{gu2023mamba} employed it to evaluate the mamba architecture. We compare the performance of SEDD and GPT-2 in selecting the correct continuation following a prompt.
Since SEDD and GPT-2 are small by current standards and are not instruction-tuned, direct prompting for the correct answer leads to poor performance. As such, their performance is evaluated based on their predictions on each possible continuations following a given prefix.
The continuations on the LAMBADA dataset are often one token long. Therefore, an answer is considered selected if it would be sampled by greedy decoding. For the other tasks (HellaSwag, PIQA, Arc and WinoGrande), the likelihood of all continuations is computed. For example, if a question has 4 possible answers, we compute the likelihood of each answer as a continuation of the question. An answer is considered selected if it has the highest likelihood among all continuations. Finally, in all cases, the accuracy is the fraction of examples where the correct continuation is the most likely.
\subsection{Generation speed}
\label{sec:timing}
\begin{table*}[t]
    \centering
    \begin{tabular}{|l|c|c|}
         \hline
         \textbf{Model} & \textbf{GPT-2} & \textbf{SEDD} \\
         \hline
         \# params & 1.3B & 1.45B \\
         \hline
         Context length & 2048 & 2048 \\
         \hline
         Positional encoding & absolute & RoPE \\
         \hline
         Num Layers & 24 & 24 \\
         \hline
         Embedding dim. & 2048 & 2048 \\
         \hline
         Num. heads  & 32 & 32 \\
         \hline
         MLP multipler & 4 & 4 \\
         \hline
         Time embedding dim. & NA & 128 \\
         \hline
    \end{tabular}
    \caption{Hyperparameters of the GPT-2 and SEDD models used to measure the generation speed.}
    \label{tab:hparams-timing}
\end{table*}
As shown in \cref{fig:sampling-speed}, sampling from a SEDD model of 1.45B parameters with 64 steps is faster than sampling from a GPT-2 model of 1.3B parameters with KV-caching. See \cref{tab:hparams-timing} for more details on the models.
\subsection{Reproducing the MAUVE results}
\label{sec:reproduce-mauve-results}
We could not reproduce the MAUVE results of \citet{lou2023discrete}. Our findings suggest that GPT-2 outperforms SEDD in conditional generation, for both small and medium models. Additionally, we observed that MAUVE is very sensitive to changes in hyperparameters. For example, simply changing the random seed could cause the score for the same text to fluctuate between 0.89 and 0.93. Hence, we use the default parameters of MAUVE with max sequence length of 100 tokens, and average the score over 5 seeds. Interestingly, MAUVE seems to correlate positively with the perplexity of the generated text, as computed using GPT-2 large. Top-k sampling is an exception. However, the low perplexity of text generated with top-k sampling might indicate repetitions rather than improved quality, since it is significantly lower than the other experiments. 

Importantly, it appears that SEDD performs worse at generating sentences shorter than it was trained for. We evaluate performance using prompts and continuations of 50 tokens each. Surprisingly, we observe that when sampling 1024 tokens and retaining the first 100, we obtain better MAUVE scores than when sampling 100 only. Since the attention mechanism of SEDD is non-causal, the later token, unused to compute MAUVE, can technically constitute a scratchpad during sampling. Investigating this surprising discrepancy is left for future work. See \cref{table:cond-quality-mauve-ppl-small,table:cond-quality-mauve-ppl-medium} for numerical results.
\end{document}